\documentclass[conference]{IEEEtran}
\usepackage{times}
\usepackage{graphicx} 
\usepackage{subfigure} 

\usepackage{amsmath}

\usepackage{algorithm}
\usepackage{algorithmic}
\newcommand{\nn}{\nonumber}

\ifCLASSINFOpdf
\else
\fi
\begin{document}
%
\title{A Deep Embedding Model for Co-occurrence Learning}


\author{\IEEEauthorblockN{Yelong Shen}
\IEEEauthorblockA{Microsoft Research\\
One Microsoft Way, Redmond, WA\\
Email: yeshen@microsoft.com}
\and
\IEEEauthorblockN{Ruoming Jin}
\IEEEauthorblockA{Kent State University\\
800 E Summit St, Kent, OH\\
Email: jin@cs.kent.edu}
\and
\IEEEauthorblockN{Jianshu Chen}
\IEEEauthorblockA{Microsoft Research\\
One Microsoft Way, Redmond, WA\\
Email: jianshuc@microsoft.com}
\and
\IEEEauthorblockN{Xiaodong He}
\IEEEauthorblockA{Microsoft Research\\
One Microsoft Way, Redmond, WA\\
Email: xiaohe@microsoft.com}
\and
\IEEEauthorblockN{Jianfeng Gao}
\IEEEauthorblockA{Microsoft Research\\
One Microsoft Way, Redmond, WA\\
Email: jfgao@microsoft.com}
\and
\IEEEauthorblockN{Li Deng}
\IEEEauthorblockA{Microsoft Research\\
One Microsoft Way, Redmond, WA\\
Email: deng@microsoft.com}
}


%


\maketitle

\begin{abstract}
Co-occurrence Data is a common and important information source in many areas, such as the word co-occurrence in the sentences, friends co-occurrence in social networks and products co-occurrence in commercial transaction data, etc, which contains rich correlation and clustering information about the items. In this paper, we study co-occurrence data using a general energy-based probabilistic model, and we analyze three different categories of energy-based model, namely, the $L_1$, $L_2$ and $L_k$ models, which are able to capture different levels of dependency in the co-occurrence data. We also discuss how several typical existing models are related to these three types of energy models, including the Fully Visible Boltzmann Machine (FVBM) ($L_2$), Matrix Factorization ($L_2$), Log-BiLinear (LBL) models ($L_2$), and the Restricted Boltzmann Machine (RBM) model ($L_k$). Then, we propose a Deep Embedding Model (DEM) (an $L_k$ model) from the energy model in a \emph{principled} manner. Furthermore, motivated by the observation that the partition function in the energy model is intractable and the fact that the major objective of modeling the co-occurrence data is to predict using the conditional probability, we apply the \emph{maximum pseudo-likelihood} method to learn DEM. In consequence, the developed model and its learning method naturally avoid the above difficulties and can be easily used to  compute the conditional probability in prediction. Interestingly, our method is equivalent to learning a special structured deep neural network using back-propagation and a special sampling strategy, which makes it scalable on large-scale datasets. Finally, in the experiments, we show that the DEM can achieve comparable or better results than state-of-the-art methods on datasets across several application domains.   
\end{abstract}


%
\IEEEpeerreviewmaketitle

\section{Introduction}
\label{introduction}
Co-occurrence data is an important and common data signal in many scenarios, for example, people co-occurrence in social network, word co-occurrence in sentences, product co-occurrence in transaction data, etc. By indicating which items appear together in each data sample, it provides rich information about the underlying correlation between different items, from which useful information can be extracted. There are several well-known machine learning models developed for analyzing co-occurrence data, e.g., topic model for bags-of-words \cite{blei2003latent}; Restricted Boltzmann Machine \cite{salakhutdinov2007restricted} and Matrix Factorization \cite{srebro2004maximum} method for collaborative filtering. These statistical models are designed for discovering the implicit or explicit hidden structure in the co-occurrence data, and the latent structures could be used for domain specific tasks.

In this paper, we study the unsupervised learning over general co-occurrence data, especially the learning of the probability distribution of the input data, which is a fundamental problem in statistics. One of the main objectives of learning a probabilistic model from co-occurrence data is to predict potentially missing items from existing items, which can be formulated as computing a conditional probability distribution. In this paper, we focus on energy-based probabilistic models, and develop a deep energy model with high capacity and efficient learning algorithms for modeling the co-occurrence data. Before that, we first systematically analyze the ability of the energy model in capturing different levels of dependency in the co-occurrence data, and we recognize three different categories of energy models, namely, Level 1 ($L_1$), Level 2 ($L_2$) and Level k ($L_k$) models. The $L_1$ models consider the components of the input vectors (aka. \emph{items}) to be independent of each other, and  joint occurrence probability of the items can be completely characterized by the popularity of each item.  The $L_2$ models assumes items occurs in data are bi-dependent with each other. The typical $L_2$ models are Ising model \cite{ravikumar2010high} and Fully Visible Boltzmann Machine (FVBM) \cite{hyvarinen2006consistency}. And the model based on $L_k$ assumption is capable of capturing any high-order (up to $k$) dependency among items. Restricted Boltzmann Machine (RBM) is an example of $L_k$ model \footnote{In \cite{le2008representational}, RBM model is proved to be a universal approximator, if the size of hidden states is exponential to the input dimension.}. However, RBM remains a shallow model with its capacity restricted by the number of hidden units. Furthermore, we also study several existing latent embedding models for co-occurrence data, especially, Log-BiLinear (LBL) word embedding model \cite{mnih2007three} and matrix factorization based linear embedding model \cite{zhang2007binary}. Both of them could be interpreted as Bayesian $L_2$ model, which are closely related to the FVBM model. 
Motivated by such the observation, we propose a Deep Embedding Model (\textbf{DEM}) with efficient learning algorithm from energy-based probabilistic models in a \emph{principled} manner for mining the co-occurrence data. \textbf{DEM} is a bottom to top hierarchical energy-based model, which incorporates both the low order and the high order item-correlation features within a unified framework. With such deep hierarchical representation of the input data, it is able to capture rich dependency information in the co-occurrence data. During the development of the model and its training algorithm, we make several important observations. First, due to the intractability of the partition function in energy models, we avoid the use of the traditional maximum-likelihood for learning our deep embedding model. Second, since our objective of modeling the co-occurrence data is to predict potentially missing items from existing items, the conditional probability distribution is the point of interest after learning. With such observations, we will show that the conditional probability distribution is indeed independent of the partition function, and is determined by the \emph{dynamic energy function} \cite{sohl2011minimum}, which is easy to compute. Moreover, such an observation naturally also points us to use the \emph{maximum pseudo-likelihood} \cite{gong1981pseudo} method to learn the deep model. Interestingly, we find that the maximum pseudo-likelihood method for learning DEM is equivalent to training a deep neural network (DNN) using (i) back-propagation, and (ii) a special sampling strategy to artificially generate the supervision signal from the co-occurrence data. The equivalent DNN has sigmoid units in all of its hidden layers and the output layer, and has an output layer that is fully connected to all its hidden layers. Therefore, the training algorithm is a discriminative method, which is efficient and scalable on large-scale datasets. Finally, in experiments, we show that \textbf{DEM} could achieve comparable or significantly better results on datasets across different domains than the state-of-the-art methods.

\textbf{Paper Organization:} {In Section 2, we provide a brief review of the related work on statistic models on co-occurrence data. In Section 3, we formally describe the Bayesian dependence framework for learning the high-dimensional binary data distribution. In Section 4, we introduce the deep embedding model and pseudo-likelihood principle for model parameter estimation, and in Section 5, we report the detailed experimental results, and finally conclude the paper in Section 6}.

\section{Related Work}
There are several proposed models in the literature for estimating the distribution of binary data. The Bayesian mixture model \cite{blei2003latent} \cite{Hofmann:1998:SMC:888741} is the most common one, which assumes the binary data to be generated from multivariate Bernoullis distribution. In \cite{frey1998graphical}, it argued that a better performance can be achieved by modeling the conditional probability on items with log-linear logistic regressors. The proposed model is named fully visible sigmoid belief networks.   While RBM proposed in \cite{hinton2006reducing}, is a universal approximator for arbitrarily data distribution. It is shown in \cite{larochelle2010tractable} that, tractable RBMs could outperform standard mixture models. Recently, a new Neural Autoregressive Distribution Estimator (NADE) is proposed in \cite{larochelle2011neural}. Experiments demonstrate that NADE could achieve significant improvement over RBMs on density estimation problem. However, the limitation of NADE is that it requires the a priori knowledge of the dependence order of the variables. Although NADE could achieve promising results for modeling data distribution, it is intractable for estimating the conditional probability of variables. Furthermore, a multi-layer neural network method is proposed for data density estimation in \cite{bengio1999modeling}. But the high model complexity (the number of free model parameters is $O(H N^2)$, where $H$ is the number of hidden neurons, and $N$ is the dimension of input data) restricts its application in practice.   
   
Dimension Reduction, i.e., \cite{Mordohai:2010:DEM:1756006.1756018} and matrix factorization i.e., \cite{Globerson:2007:EEC:1314498.1314572} are two common types of embedding techniques. However, both of the two approaches focus on learning low-dimensional representation of objects while reserving their pair-wise distances. However, the co-occurrence data may have the high-order dependence (we will discuss it in the following section); Therefore, instead of reserve the distance between one object to another, the Deep Embedding model would capture the high-order dependence, i.e., correlation between multiple-objects and another one.
   
The Deep Embedding Model (DEM) proposed in this paper is derived as a model for estimating the data distribution. We evaluate the performance of DEM on the missing item prediction task, which will show that the proposed DEM significantly outperforms most of the existing models.    

DEM is also closely related to the autoencoder \cite{vincent2008extracting} models, which contains two components: encoder and decoder. The encoder maps the input data to hidden states, while the decoder reconstructs the input data from the hidden states. There are also some studies to connect denoising auto encoder to generative learning \cite{bengio2013generalized,vincent2011connection,swersky2011autoencoders}. Indeed, DEM could be also viewed as an special case of denoising autoencoder. In encoder phase, the input data is corrupted by randomly dropping one element, and is then fed into the encoder function to generate the hierarchical latent embedding vectors. Then, in the decoding phase, the missing items are reconstructed from latent vectors afterwards.

\section{Co-occurrence Data Modeling}  
\label{coocc}
In this section, we first introduce the basic notation of the paper and then present the Bayesian dependency framework for analyzing the existing models. 

Let $V$ denote the set of the co-occurrence data, which contain $N$-dimensional binary vectors $v \in \{0,1\}^N$, where $N$ is the total number of items. Specifically, the value of the $n$-th entry of the vector $v$ is equal to one if the corresponding item occurs, and it is equal to zero if it does not occur. For example, in word co-occurrence data, $N$ denotes the vocabulary size, and the values of the entries in vector $v$ denote whether the corresponding words appear in the current sentence.

The fundamental statistical problem for co-occurrence learning can be formulated as estimating the probability mass function (pmf), $p_{\theta}(v), v \in \{0,1\}^N$, from the observation dataset, $V$. A straight-forward method for pmf estimation is to count the frequency of occurrence of the $v$ in the entire corpus $V$, given $V$ contains infinite i.i.d samples. However, it is unrealistic in practice because it requires us to learn a huge table of $2^N$ entries, where $N$ can be as large as tens of thousands in many applications. Therefore, a practically feasible method should balance the model complexity and capability for co-occurrence data modeling. Throughout the paper, we consider the probability mass function $p_{\theta}(v)$ that can be expressed by the following general parametric form:
	\begin{align}		
		p_{\theta}(v)	=	\frac{1}{Z} e^{-E_{\theta}(v)},
		\quad 
		v \in  \{0, 1\}^N
		\label{Equ:CooccurrenceModel:GeneralEnergyModel}
	\end{align}
where $E_{\theta}(v)$ is the energy function on data $v$ with parameter $\theta$, and $Z$ is the partition function that normalizes $p_{\theta}(v)$ so that it sums up to one, which is a function of $\theta$. In the following subsections, we introduce three Bayesian dependence assumptions, namely, $L_1$, $L_2$ and $L_k$ on the model \eqref{Equ:CooccurrenceModel:GeneralEnergyModel}, where the energy function $E_{\theta}(v)$ would assume different forms under different assumptions. Within this framework, we will show that several popular statistical models fall into different categories (special cases) of the above framework, and we will also explain how different types of models are able to trade model capacity with model complexity. Moreover, the Bayesian dependence framework would further motivate us to develop a deep embedding model for modeling the co-occurrence data, which will be discussed in Section \ref{Sec:DeepEmbedModel}.




\subsection{Bayesian $L_1$ Dependence Assumption}
We first consider the $L_1$ Bayesian Dependence Assumption, where the items in co-occurrence data are assumed to be independent of each other so that the probability mass function of $v$ can be factorized into the following product form:
\begin{equation}
\label{l1d}
\textbf{$ p_{\theta}(v) = \prod_{i \in I_{v}}{p(i)} \prod_{i \notin I_v }{(1-p(i))} $}
\end{equation}
where $I_{v}$ denotes the set of the items occurred in $v$, and $p(i)$ is the occurrence probability of the $i$-th item. Note that, in this case, the joint probability mass function $p_{\theta}(v)$ is factored into the product of the marginal probabilities of the entries of the vector $v$. The pmf in \eqref{l1d} could be further rewritten in the parametric form \eqref{Equ:CooccurrenceModel:GeneralEnergyModel} with the energy function in this case being
	\begin{align}
		E_{\theta}^{L_1}(v) = b^T v = \sum_{i \in I_v} b_{i}
		\nn
	\end{align}
where $b_i = - \ln p(i) + \ln (1-p(i)) $ is the negative log-likelihood ratio for the $i$-th item.
\subsection{Bayesian $L_2$ Dependence Assumption}

Likewise, for the Bayesian $L_2$ dependence, the energy function $E_{\theta}(v)$ in \eqref{Equ:CooccurrenceModel:GeneralEnergyModel} assumes the following form:
	\begin{align}
		E_{\theta}^{L_2}(v) = v^T W v + b^T v
		\label{Equ:CoocurrenceModel:L2Energy}
	\end{align}
where $W$ is a $N \times N$ symmetric matrix with zero diagonal entries. The energy function \eqref{Equ:CoocurrenceModel:L2Energy} could also be written in the following equivalent form:
	\begin{equation}
		\label{energyl2d}
		\textbf{$ E^{L_2}_{\theta}(v) = \sum_{i \in I_v} b_{i} + \sum_{i , j \in I_v (i \neq j)} W_{ij} $}
	\end{equation}
One typical model with $L_2$ assumption is Markov Random Field Model (or Fully Visible Boltzmann Machine model (FVBM) \cite{hyvarinen2006consistency}, or Ising Model \cite{ravikumar2010high} ), which is widely used in image modeling \cite{geman1986markov}.

\subsection{Bayesian $L_k$ Dependence Assumption}

The Bayesian $L_k$ dependence assumption is proposed to model any high-order correlations among items in co-occurrence samples. Thus, we extend the classical $L_2$ FVBM model with $L_k$ FVBM. The new energy function for $L_k$ FVBM could be given as follows:
\begin{align}
\label{energylkd}
E_{\theta}^{L_k}(v)	&= 		\sum_{i \in I_v} b_{i} 
							+ \!\!\!\!\!\!\sum_{i , j \in I_v (i \neq j)} 
							\!\!\!\!\!\!\!\!
							W_{ij} 
							+
							... 
							+
							\!\!\!\!\!\!\!\!\!\!\!\! 
							\sum_{i , j,..,k \in I_v (i \neq j .. \neq k)} 
							\!\!\!\!\!\!\!\!\!\!\!\!\!\!\! 
							W_{ij..k}
\end{align}
Note that, as $k$ increases, the above energy function is able to capture high-order correlation structures, and the model complexity also grows exponentially with $k$.


\subsection{Conditional Probability Estimation}
\label{Sec:CoocurrenceModel:CondProbEst}

So far we have introduced the energy-based probabilistic model for co-occurrence data and its particular forms in modeling different levels of dependency, i.e., $L_1$, $L_2$ and $L_k$ models. The classical approach for learning the model parameters of such an energy-based model is the maximum likelihood (ML) method. However, the major challenge of using the ML-based method is the difficulty of evaluating the partition function $Z$ and its gradient (as a function of $\theta$) in the energy model \eqref{Equ:CooccurrenceModel:GeneralEnergyModel}. Nevertheless, in many practical problems, the purpose of learning the probability distribution of the input (co-occurrence) data is to predict a potentially missing item given a set of existing items. That is, the potential problem is to find the probability of certain elements of the vector $v$ given the other elements of $v$.  For example, in the item recommendation task, the objective is to recommend new items that a customer may potentially purchase given the purchasing history of the customer. In these problems, the conditional probability of the potentially missing items given the existing items is the major point of interest. As we will proceed to show, learning an energy model that is satisfactory for prediction using its associated conditional probability does not require estimating the partition function in \eqref{Equ:CooccurrenceModel:GeneralEnergyModel}. In fact, we now show that it is actually convenient to compute the conditional probability from the energy model \eqref{Equ:CooccurrenceModel:GeneralEnergyModel}. Specifically, the conditional probability can be computed from the energy function via the following steps:							
	\begin{align}
	\ln p_{\theta}(v_t = 1 | v_{(-t)}) 	&=
							\ln \frac{p_{\theta}(v_t = 1, v_{(-t)})}
							{p_{\theta}(v_t = 1, v_{(-t)}) + p_{\theta}(v_t = 0, v_{(-t)})}
							\nn\\
							&=
							\ln \frac{p_{\theta}(v_{(+t)})}{p_{\theta}(v_{(+t)}) + p_{\theta}(v_{(-t)})}
							\nn\\
							&=
							\ln \frac{e^{- E_{\theta}(v_{(+t)})}} {e^{-E_{\theta}(v_{(+t)})} + e^{-E_{\theta}(v_{(-t)})} }
							\nn\\
						&=
							\ln \frac{1}{1+ \exp\left\{{E_{\theta}(v_{(+t)}) - E_{\theta}(v_{(-t)}) }\right\}}
							\nn\\
						&= \ln \sigma\left(E_{\theta}(v_{(-t)}) - E_{\theta}(v_{(+t)}) \right)
	\label{cpe}
	\end{align}
where $v_{(-t)} \in \{0,1\}^N$ is the input vector indicates the existing items (with $t$-th entry being zero), $v_{(+t)} \in \{0,1\}^N$ is an $N$-dimensional vector (with the $t$-th entry being one and all other entries equal to $v_{(-t)}$); $\sigma(\cdot)$ is the logistic function defined as $\sigma(x) = 1/(1+e^{-x})$. Since $v_{(+t)}$ is only one bit different from $v_{(-t)}$, we define the dynamic energy function \footnote{Jascha Sohl-Dickstein et al. first introduced the concept of dynamic energy in minimum probability flow method \cite{sohl2011minimum}.} as 
	\begin{align}
		F_{\theta}(t, v) = E_{\theta}(v_{(+t)}) - E_{\theta}(v)
		\label{Equ:DynamicEnergy_def}
	\end{align}
where let $v$ equals to $v_{(-t)}$ for notation simplification. As a result, the log conditional probability can be written as
	\begin{align}
		\ln p_{\theta}(v_t = 1 | v_{(-t)}) 	&=
									\ln \sigma\left( F_{\theta}(t, v) \right)
		\label{Equ:LogCondProb_DynamicEnergy}
	\end{align}
Note from \eqref{Equ:LogCondProb_DynamicEnergy} that the conditional probability $p_{\theta}(v_t = 1 | v_{(-t)})$, which is of interest in practice, no longer depends on the partition function, but only on the dynamic energy function $F_{\theta}(t,v)$. Therefore, from now on, we only need to study the specific form of the $F_{\theta}(t,v)$ for different $L_1$, $L_2$ and $L_k$ models, which can be computed as
\begin{align}
\label{l12kdef}
F_{\theta}^{L_1}(t, v) &= b_t \\
F_{\theta}^{L_2}(t, v) &= b_t + \sum_{i \in I_{v}} W_{it} \\
F_{\theta}^{L_k}(t, v) &= b_t + \sum_{i \in I_{v}} W_{it} + \cdots + 
							\!\!\!\!\!\!\!\!\!\!\!\! 
							\sum_{i,..,k \in I_v (i \neq .. \neq k)} 
							\!\!\!\!\!\!\!\!\!\!\!\!\!\!\! 
							W_{i..kt}							
\end{align}
where $F_{\theta}^{L_1}(t, v)$ is a constant function for any given $t$; $F_{\theta}^{L_2}(t, v)$ is a linear function; and $F_{\theta}^{L_k}(t, v)$ is a nonlinear function of the variable $v$.

\subsection{Relation to Several Existing Models}

We now briefly introduce the relation of our $L_1$, $L_2$ and $L_k$ formulation to several typical existing models for co-occurrence data modeling.

{\bf Log-Bilinear (LBL)  Embedding Model:} Mnih and Hinton et al. \cite{mnih2007three} introduce a neural language model which uses a log-bilinear energy function to model the word contexts. In its Log-Bilinear model, the posterior probability of a word given the context words is given by \cite{maas2010probabilistic} \footnote{The original Log-Bilinear model contains the transform matrix $C$ for modeling word position information. i.e., conditional probability for next word: $\log p_{\theta}(t | v) \propto {\phi^T_{t} \!\sum_{i \in I_{v}} \!  \phi_{i} C_i } $. We remove the transform matrix $C$ since the position information is assumed to be not available in co-occurrence data. }:
\begin{align}
\label{lblembed}
\log p_{\theta}(t | v) \propto 
	{\phi^T_{t} \!\sum_{i \in I_{v}} \!  \phi_{i} } 
	= \phi^T_{t} \Phi v 
\end{align}
where $\phi_t$ is the vector representation of word $t$, $\Phi$ is the word embedding lookup table. $\phi_t$ is the $t$-th row of $\Phi$. 
As we can see, the formulation \eqref{lblembed} is a linear function over $v$ for any given $t$. Thus, LBL embedding model, to some extent, can be interpreted as an $L_2$ dependence model.

{\bf Matrix Factorization:} Matrix Factorization based approaches are probably the most common latent embedding models for co-occurrence data. The maximum margin matrix factorization (MMMF) model \cite{srebro2004maximum} learns the latent embedding of items based on the following objective function:
\begin{align}
\label{mmmf}
(\Phi, Z) = \arg\min_{\Phi, Z} \sum_{v^i \in V} (v^i - \Phi z^i)^2 + \lambda |\Phi|^2 + \beta |Z|^2
\end{align} 
where $v^i$ denotes the $i$-th data sample for training, $z^i$ is the latent representation for the $i$-th sample, and $\Phi$ is the item embedding matrix.

When predicting the scores of missing items from observation $v$, MMMF first estimates the hidden vector via
	\begin{align}
		z = \arg\min_z (v - \Phi z)^2 + \beta |z|^2 = (\Phi^T \Phi + \beta)^{-1} \Phi^T v
	\end{align}
and then the score function of the missing item $t$ given $v$ could be computed as
\begin{align}
\label{mmmfpred}
S(t; v) = \phi_t^T (\Phi^T \Phi + \beta)^{-1} \Phi^T v
\end{align}
where $\phi_t$ is the vector representation of item $t$, it is the $t$-th row of matrix $\Phi$.
The formulation \eqref{mmmfpred} is very similar to that of \eqref{lblembed}. Therefore, MMMF model is also related to L2 dependence model.


{\bf Restricted Boltzmann Machine (RBM):} the Restricted Boltzmann machine is a classical model for modeling data distribution. Early theoretical studies show that the RBM can be a universal function approximator. It could learn arbitrarily data distributions if the size of hidden states is exponential in its input dimension \cite{le2008representational}. Typically, an RBM is expressed in:
\begin{align}
\label{rbm}
p_{\theta}(v) = \sum_{h} p(v,h) 
			= \frac{1}{Z} \sum_{h} e^{-(h^TWv + c^T h + b^T v)} 
\end{align}
By integrating out the hidden variables in \eqref{rbm}, we could obtain its energy function on $v$:
\begin{align}
\label{rbmenergy}
E_{\theta}(v) = b^T v + \sum_{h=1}^{H} \ln ( 1 + e^{w_h v + c_h})
\end{align}
where $H$ is the number of hidden states, the term $\ln ( 1 + e^{x})$ is  the soft-plus function. It can be  considered as a smoothed rectified function. In \cite{le2008representational}, it is proved that soft-plus function can approximate any high-order boolean function. By allowing the number of hidden states be exponential to the item numbers, the energy function in \eqref{rbmenergy} could approximate the $L_k$ energy function \ref{energylkd}. Therefore, RBM can be interpreted as an $L_k$ dependence assumption.

\section{Deep Embedding Model} 
\label{Sec:DeepEmbedModel}
In this section, we present the Deep Embedding Model (DEM) for co-occurrence data modeling. As we discussed in the previous section, many classical embedding models only capture $L_2$ dependency, and RBM, although being an $L_k$ model, has its capability bounded by the number of hidden states. Motivated by the above observation, we propose a deep hierarchical structured model that is able to capture the low-order item dependency at the bottom layer, and the high-order dependency at the top layer.  As we discussed in section \ref{Sec:CoocurrenceModel:CondProbEst}, our objective is to learn an energy model that allows us to perform satisfactory prediction using its associated conditional probability $p_{\theta}(v_t=1 | v_{(-t)})$ instead of the original $p_{\theta}(v)$. And recall from \eqref{Equ:LogCondProb_DynamicEnergy} that the conditional probability is determined only by the dynamic energy function $F_{\theta}(t,v)$ and is independent of the partition function. We first propose a deep hierarchical energy model by giving its dynamic energy function, and then show how to learn the deep model efficiently. 

The dynamic energy function for the deep embedding model is given by
	\begin{align}
	\label{energyflowdem}
	F^{DEM}_{\theta}(t, v) 
		&= b_t + \! \sum_{ i \in I_v } R^{0}_{it}  \!+ \! {R^1_t} {h_1} \!+ \!
	{R^2_t} {h_2} + ... + {R^k_t} {h_k}  
	\end{align}
where $\{h_1, h_2, ..., h_k \}$ are the hidden variables computed according to a feed-forward multi-layer neural network:
\begin{align}
\label{multinn}
&h_1 = \sigma({W^1 v + B^1}) \\
&h_i = \sigma({W^i h_{i-1} + B^i} )  &  i=2,...,k
\end{align}
where $\sigma(\cdot)$ is the logistic (sigmoid) function; $\{ (W^i, B^i)_{i=1,..,k}\}$ are the model weights in multi-layer neural networks. In the expression \eqref{energyflowdem}, there is a set of hierarchical structured embedding vectors $\{R^1_t, R^2_t, ..., R^k_t\}$ assigned to each item $t$, where the inner product between the hidden variables $h_i$ and $R^i_t$ could approximate any weighted high-order boolean functions on $v$. Therefore, the proposed DEM could capture $L_k$ dependency. Notice that we also keep the terms corresponding to the $L_1$ and $L_2$ dependency in the dynamic energy function of DEM, which makes the model more adaptive to different data distribution. 

As we discussed earlier, due to the difficulty of handling the partition function in the energy model and the fact that we only need a conditional probability in our prediction tasks, we avoid the use of the traditional maximum likelihood principle \cite{johansen1990maximum} for modeling the co-occurrence data, which seeks to solve the following optimization problem:
\begin{align}
\label{mlikelihood}
\theta^* = \arg\max_{\theta} \sum_{v \in V} \ln p_{\theta}(v)
\end{align}
For the same reason, we also present our deep embedding model by giving its dynamic energy function directly, which can be used for computing the conditional probability easily.  Furthermore, in the paper, we will use an alternative approach, named \emph{maximum pseudo-likelihood principle}  \cite{gong1981pseudo} for learning the model parameters of DEM, which seeks to maximize the conditional probability function:
\begin{align}
\label{pmlikelihood}
\theta^* = \arg\max_{\theta} \sum_{v \in V} \sum_{t=1}^N \ln p_{\theta} (v_{t} | v_{(-t)}) 
\end{align}
where $v_{(-t)}$ is the data sample $v$ with the $t$-th entry missing \footnote{In the following sections, $v$ and $v_{(-t)}$ are different. They could be equal to each other when $v_t = 0$.}, and $v_t \in \{ 0, 1 \}$ is the $t$-th entry of $v$ .  By substituting \eqref{Equ:LogCondProb_DynamicEnergy} into \eqref{pmlikelihood}, we obtain
\begin{align}
\theta^* 	&=  	\arg\max_{\theta} \sum_{v \in V} \sum_{t=1}^N \ln 
				\sigma( F_{\theta}(t,v) )
			\nn\\
		&=
			\arg\max_{\theta} \sum_{v \in V} \sum_{t=1}^N \ln 
			\sigma( E_{\theta}( v_{(\hat{t})} ) - E_{\theta}(v) )
\label{newpmlikelihood}
\end{align} 
where $v_{(\hat{t})}$ is the neighbor of $v$ (with the $t$-th entry flipped from $v_t$ and all other entries equal to $v$, which has a unit Hamming distance from $v$). Note that, expression \ref{newpmlikelihood} can be further written as
\begin{align}
\label{objectfunc}
\theta^* &=  \arg\max_{\theta} \sum_{v \in V} \!\!
		\left[ \sum_{t \in I_v} \ln \sigma( - \! F_{\theta}(t, v_{(-t)}  ) )
		+ \!\!\!
		\sum_{t \notin I_v} \ln \sigma(F_{\theta}(t, v) )
		\right]
\end{align}

\begin{figure}[t]
\begin{center}
\centerline{\includegraphics[width=\columnwidth]{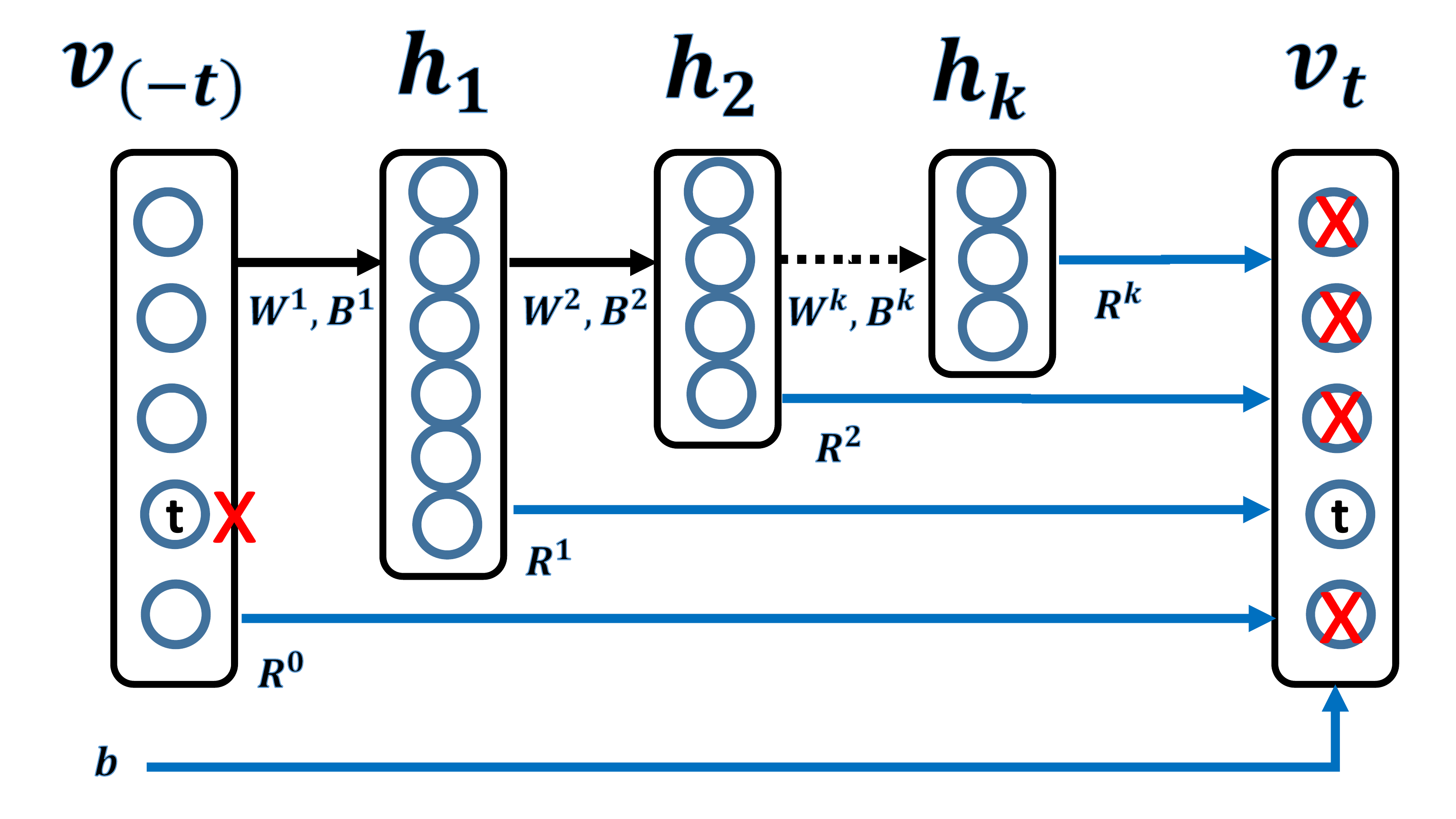}}
\caption{Illustration of the computation of the dynamic energy function in Deep Embedding Model, where the red cross means a missing item.}
\label{dem-figure}
\end{center}
\end{figure}

From \eqref{objectfunc} and Figure \ref{dem-figure}, we note that the maximum pseudo-likelihood optimization of our deep energy model is equivalent to train a deep feed-forward neural network with the following special structure:
	\begin{itemize}
		\item
		The nonlinearity of the hidden units is the sigmoid function.
		
		\item
		The output units are fully connected to all the hidden units.
		
		\item
		The nonlinearity of the output units is also the sigmoid function.
	\end{itemize}
Furthermore, the training method is performing back-propagation over such a special deep neural network (DNN). However, the method is also different form the traditional back-propagation method in its choice of the supervision signal. The traditional back-propagation method usually uses human labeled targets as its supervision signal. In the co-occurrence data modeling, there is no such supervision signal. Instead, we use a special sampling strategy to create an artificial supervision signal by flipping the input data at one element for each sample, and the algorithm performs discriminative training for such an unsupervised learning problem. Interestingly, the maximum pseudo-likelihood learning strategy for our proposed deep energy model is equivalent to discriminatively training the special DNN in Figure \ref{dem-figure} using back-propagation and a special sampling strategy. In the next subsection, we will explain the details of the training algorithm.

\subsection{Model Parameter Estimation}  

To maximize the objective function of DEM in \eqref{objectfunc}, we apply the stochastic gradient descent method to update model parameters for each data sample, $v$. We omit the details of gradient derivation from the objective function. The following updating rules are applied :

First, randomly select a element $v_t$ from $v$; If $t \in I_v$, then $v_t = 1$, otherwise $v_t = 0$. Second, compute the $\Delta(v,t)$:
\[
   \Delta(v,t) =
   \begin{cases}
        \sigma( F_{\theta}(t, v_{(-t)})), & \text{if } t \in I_v  \\
        1 - \sigma( F_{\theta}(t, v)), & \text{otherwises }.
   \end{cases}
\]
\textbf{Update $b$: }
\begin{align}
\label{updateb}
\Delta b_t = \Delta(v,t)
\end{align}
\textbf{Update $R^0$ :}
\begin{align}
\label{updater0}
\Delta R^0_{it} = \Delta(v,t) && i \in I_v (i \neq t)
\end{align}
\textbf{Update $R^i$, $W^i$ and $B^i$ : }
\begin{align}
\label{updaterk}
&\Delta R^i_{t} = \Delta(v,t) h_i \\
&\Delta W^i = ((\Delta(v,t) R^i_t + L_i) \circ h_i \circ ( 1 - h_i)) h_{i-1}^T  \\
&\Delta B_i = ((\Delta(v,t) R^i_t + L_i) \circ h_i \circ ( 1 - h_i))
\end{align}
where $h_0$ indicates $v_{(-t)}$ if $t \in I_v$; otherwise $v$; $\{L_i\}$ can be given as follows:

In the details of implementation, we do not enumerate all the $t \notin I_v$, but sample a fix number ($T$) of samples to speed up the training process. Algorithm \ref{alg:demalg} describes details of applying stochastic gradient descent method for training the Deep Embedding Model.


\begin{algorithm}[t]
   \caption{SGD for training Deep Embedding Model}
   \label{alg:demalg}
\begin{algorithmic}
   \STATE {\bfseries Input:} Data $v$, DEM model, Negative Sample Number $T$ 
   \STATE {\bfseries Output:} Updated DEM model
   \FOR  {Select $t$ from $I_v$, $t \in I_v$}
   \STATE Calculate the Dynamic Energy Function $F_{\theta}(t, v_{(-t)})$ in \eqref{energyflowdem}
   \STATE Calculate $\Delta(v,t) = \sigma( F_{\theta}(t, v_{(-t)}))$
   \STATE Update model parameters by \eqref{updateb} - \eqref{updaterk}
   \ENDFOR
   \FOR {$i=1$ {\bfseries to} $T$}
   \STATE Randomly select $t \notin I_v$   
   \STATE Calculate the Dynamic Energy Function $F_{\theta}(t, v)$ in \ref{energyflowdem}
   \STATE Calculate $\Delta(v,t) = 1 - \sigma( F_{\theta}(t, v))$
   \STATE Update model parameters by \eqref{updateb} - \eqref{updaterk}
   \ENDFOR
\end{algorithmic}
\end{algorithm}

\section{Experiment}
In this section, we validate the effectiveness of the Deep Embedding Model (DEM)  empirically on several real world datasets. The datasets are categorized into three domains: Social networks, Product Co-Purchasing Data and Online Rating Data. We first introduce details of our experiment datasets.

\textbf{Social networks :} The social networks are collected from two sources: {Epinion} \footnote {Epinion network http://snap.stanford.edu/data/soc-Epinions1.html}, {Slashdot } \footnote{Slashdot network https://snap.stanford.edu/data/soc-Slashdot0811.html}; Both of them are directed graphs. The user in social networks has an unique uid. The social connections of the user is represented as a binary sparse vector, which contains the friends-occurrence information. In experiments, users in social network datasets are divided into five cross folders, each folder contains $80$ percent users for training, and $20$ percent users for testing. For the user in test set, it will randomly remove one of her/his connections to others. Statistic models will predict the missing edge according to the existing connections. Epinion dataset contains $75,879$ users and $508,837$ connections; Slashdot dataset contains $77,360$ users and $905,468$ connections;   

\textbf{Product Co-Purchasing Data :} Product co-purchasing datasets are collected from a anonymous Belgian Retail store \footnote{Retail dataset http://fimi.ua.ac.be/data/retail.data}. The transaction sets are divided into five cross folders, each folder contains $80$ percent users for training, and $20$ percent users for testing. For each transaction record in test set, it will randomly remove one item from the list. Model performance is measured by the number of missing items being correctly recovered. In the Retail dataset, it contains $15,664$ unique items, $87,163$ transaction records and $638,302$ purchasing items.
  
\textbf{Online Rating Data :} Online Rating datasets contains two datasets --- MovieLen10M and Jester. MovieLen10M \footnote{MovieLen10M http://grouplens.org/datasets/movielens/} is the movie rating data set with ratings ranging from 1 to 5. Jester \footnote{Jester dataset www.ieor.berkeley.edu/~goldberg/jester-data/} is an online joke recommender system. Users could rate jokes with continuous ratings ranging from -10 to 10. Users in rating dataset can be represented as sparse rating vectors. In our experiment setting, we transform the real-value ratings into the binary value by placing rating threshold, i.e., ratings equal or larger than four will be treated as one, otherwise zero in MovieLen10M dataset. In Jester, the rating threshold is zero. Jester dataset contains $101$ unique jokes, $24,944$ users, and $756,148$ ratings above zero. MovieLen10M contains $10,104$ unique movies, $69,765$ users and $3,507,735$ ratings equal or larger than four. Both the two datasets are divided into five cross folders, each folder contains $80$ percent users for training, and $20$ percent users for testing. 

\subsection{Evaluation}
In the experimental study, we make use of the missing item prediction task for evaluating model performance. All the data sets are divided into five cross folders. Records in test sets are represented as an binary sparse vector with one of its nonzero element missed. For each test record $v$, we use $g_v$ to denote the ground truth of the missing item index, $P_K(v)$ denote the predicting TopK item index list. \textbf{TopK Accuracy} is used as the main evaluation metric in experiments. The formal definition of \textbf{TopK Accuracy} is given as follows:
\begin{align}
\label{topkmeasure}
Top@K Acc = \frac{1}{|T|} {\sum_{v \in T} I(g_v \in P_K(v))}
\end{align}
where $T$ is the whole test set, $I(x)$ is the boolean indicator function; If $x$ is true, $I(x) = 1$; otherwise $I(x) = 0$. In experiments, \textbf{Top@1 Acc} and \textbf{Top@10 Acc} are two key indicators for model comparison. 

\subsection{Experiment Results}
In this section, we report the performance of proposed Deep Embedding Model (DEM) compared with other state-of-the-art baselines. Specifically, the following baseline methods are compared: 

\textbf{Co-Visiting Graph (CVG)} \cite{das2007google}: Co-Visiting Graph method computes the item co-occurrence graph; where the weighted edge between two items is the number of times the items co-occur; In prediction phase, \textbf{CVG} scores the candidate item by summing all the edge weights linked from existing items. 

\textbf{Normalized CVG (Norm CVG)}: \textbf{Norm CVG} is an variant of \textbf{CVG} method; where the edge weight in \textbf{Norm CVG} is normalized by the frequency of items.
   
\textbf{Local Random Walk (LRW)} \cite{yildirim2008random}: \textbf{LRW} computes the similarity between a pair of items by simulating the probability of a random walker revisiting from the initial item to the target item. \textbf{LRW} method performs random walk algorithm based on the co-visiting graph, it could be alleviating the sparsity problem in the graph. In experiments, the number of steps in random walk algorithm are varied from 1 to 4; The results reported are based on the parameter configurations which produce the best results.

\textbf{Latent Dirichlet Allocation (LDA)} \cite{blei2003latent}: \textbf{LDA} model can be viewed as an variant of matrix factorization approach, where the items co-occurrence information is assumed to be generated by latent topics. In prediction phase, \textbf{LDA} estimates the latent topic distribution given existing observed items, and it generates the most probable missing items according to topic distribution. The number of topics in \textbf{LDA} model is varied from $32$ to $512$ in experiments.

\textbf{Restricted Boltzmann Machine (RBM)}\cite{salakhutdinov2007restricted} : \textbf{RBM} is an general density estimation model, which could be naturally used for missing prediction task \cite{salakhutdinov2007restricted}. In the experiments, the number of hidden states in RBM model is varied from $32$ to $512$.

\textbf{LogBilinear (LBL) Model} \cite{mnih2007three} : \textbf{LBL} is first proposed for language modeling task \cite{mnih2007three}. In our experiments, item position information is not available. Therefore, a simpler version of \textbf{LBL} model is implemented by removing the position variables. The number of embedding dimension for \textbf{LBL} is varied from $32$ to $512$ in experiments.
 
\textbf{Fully Visible Boltzmann Machine (FVBM)}\cite{hinton1986learning} : \textbf{FVBM} is an type of Markov Random Field Model as described in section 2.

\textbf{Denoising AutoEncoder (DAE)} \cite{bengio2013generalized}.

\textbf{Deep Embedding Model (DEM)} : \textbf{DEM} could be configured with different number of hidden layers and different number of hidden states. In experiments, we select the number of hidden states varied from $8$ to $512$, and the number of hidden layers from $1$ to $3$.

\begin{table*}[t]
\caption{Top@1 and Top@10 Prediction Accuracy on MovieLen10M(500) and Jester(101) Dataset.
 Superscripts $\alpha$, $\beta$, $\gamma$ and $\delta$ indicate statistically significant improvements $(p < 0.01)$ over DAE, FVBM, LBL and RBM}
\label{topkratingRecommendation_crossfolder_small}
\begin{center}
\begin{sc}
\begin{tabular}{c|lll|lll}
\hline
\multicolumn{1}{c}{} & & & & \\[\dimexpr-\normalbaselineskip-\arrayrulewidth]%

\textbf{Models} & \multicolumn{3}{c|}{MovieLen10M (Top500 Movies)} & \multicolumn{3}{c}{Jester (101 Jokes)} \\
				 	& Top@1 	& Top@10	& Run Time	&	Top@1	& 	Top@10 & Run Time	\\ 
\hline 
CVG   		& $4.26 \pm 0.23 $  & $19.56 \pm 0.21$	& $\approx$ 10 sec & $16.59 \pm 0.20$ & $59.70 \pm 0.34 $ & $\approx$ 2 sec \\ [2pt]
\hline 
NormCVG   	& $4.73 \pm 0.23 $  & $21.02 \pm 0.25$ & $\approx$ 10 sec  & $16.65 \pm 0.25$ & $59.98 \pm 0.36 $ & $\approx$ 2 sec \\[2pt]
\hline
LRW   		& $4.40 \pm 0.24 $  & $20.28 \pm 0.33$ & $\approx$ 50 sec & $16.57 \pm 0.22$ & $59.98 \pm 0.39 $ & $\approx$ 10 sec \\[2pt]
\hline
LDA    	    & $6.70 \pm 0.35$  & $30.11 \pm 0.41$ & $\approx$ 1000 sec & $15.88 \pm 0.30$ & $62.88 \pm 0.41 $ &  $\approx$ 100 sec \\[2pt]
\hline
RBM		    & $10.52 \pm 0.33$  & $39.40 \pm 0.63$ & $\approx$ 500 sec & $19.66 \pm 0.29 $ & $68.95 \pm 0.67$ & $\approx$ 50 sec \\[2pt]
\hline
LBL         & $10.42 \pm 0.33$  & $38.49 \pm 0.44$ & $\approx$ 300 sec & $\textbf{20.21} \pm \textbf{0.27} $ & $\textbf{69.46} \pm \textbf{0.59}$ & $\approx$ 10 sec \\[2pt]
\hline
FVBM   		& $10.77 \pm 0.35$  & $39.34 \pm 0.48$ &  $\approx$ 400 sec & $\textbf{20.35} \pm \textbf{0.28} $ & $\textbf{69.18} \pm \textbf{0.42}$ & $\approx$ 10 sec \\[2pt]
\hline
DAE			& $10.50 \pm 0.34$  & $39.41 \pm 0.47$ & $\approx$ 200 sec & $19.35 \pm 0.35 $ & $68.16 \pm 0.77$ & $\approx$ 10 sec \\[2pt]
\hline
\textbf{DEM}& $\textbf{11.32} \pm \textbf{0.42}$ $^{\alpha\beta\gamma\delta}$  & $\textbf{41.33} \pm  \textbf{0.75}$ $^{\alpha\beta\gamma\delta}$ & $\approx$ 400 sec & $\textbf{20.56} \pm \textbf{0.23}$ $^{\alpha\delta}$ & $\textbf{69.46} \pm \textbf{0.66}$ $^{\alpha\delta}$ & $\approx$ 10 sec \\[3pt]
\hline
\end{tabular}
\end{sc}
\end{center}
\end{table*}

The experiment environment is built upon machine Inter Xeon CPU 2.60 (2 Processors) plus four Tesla K40m GPUs. Except CVG, NormCVG, LRW and LDA methods, all other approaches run on GPU. 

In the Table \ref{topkratingRecommendation_crossfolder_small}, we provide a detailed comparison of these nine approaches in terms of Top@1 and Top@10 prediction accuracy on MovieLen10M (Top500 Movies) and Jester datasets. Proposed \textbf{DEM} method shows significant improvements over all baselines on MovieLen10M (Top500) dataset. On Jester dataset, \textbf{DEM} significant outperforms other baselines except \textbf{LBL} and \textbf{FVBM}. The running time includes both training time and prediction time.

\begin{table*}[t]
\caption{Top@1 and Top@10 Prediction Accuracy on MovieLen10M (10,269) and Retail(16,469) Dataset.  Superscripts $\alpha$, $\beta$, $\gamma$ and $\delta$ indicate statistically significant improvements $(p < 0.01)$ over DAE, FVBM, LBL and RBM}
\label{topkratingRecommendation_crossfolder_large}
\begin{center}
\begin{sc}
\begin{tabular}{c|lll|lll}
\hline
\multicolumn{1}{c}{} & & & & \\[\dimexpr-\normalbaselineskip-\arrayrulewidth]%

\textbf{Models} & \multicolumn{3}{c|}{MovieLen10M (10,269 Movies)} & \multicolumn{3}{c}{Retail (16,469 Items)} \\
				 	& Top@1 	& Top@10	& Run Time	&	Top@1	& 	Top@10 & Run Time	\\ 
\hline 
CVG   		& $3.24 \pm 0.12 $  & $14.34 \pm 0.15$	& $\approx$ 10 sec & $13.48 \pm 0.17$ & $25.30 \pm 0.27 $ & $\approx$ 10 sec \\ [2pt]
\hline 
NormCVG   	& $3.74 \pm 0.13 $  & $16.01 \pm 0.17$ & $\approx$ 10 sec  & $13.92 \pm 0.10$ & $28.01 \pm 0.36 $ & $\approx$ 10 sec \\[2pt]
\hline
LRW   		& $3.54 \pm 0.14 $  & $15.51 \pm 0.08$ & $\approx$ 1800 sec & $13.92 \pm 0.08$ & $27.56 \pm 0.34 $ & $\approx$ 200 sec \\[2pt]
\hline
LDA    	    & $4.15 \pm 0.13$  & $18.95 \pm 0.06$ & $\approx$ 2600 sec & $13.30 \pm 0.16$ & $24.76 \pm 0.32 $ &  $\approx$ 1300 sec \\[2pt]
\hline
RBM		    & $4.69 \pm 0.17 $  & $20.80 \pm 0.02$ & $\approx$ 1800 sec & $12.74 \pm 0.29 $ & $23.72 \pm 0.37$ & $\approx$ 800 sec \\[2pt]
\hline
LBL         & $6.67 \pm 0.12$  & $26.45 \pm 0.33$ & $\approx$ 700 sec & $\textbf{15.05} \pm \textbf{0.20} $ & $\textbf{26.00} \pm \textbf{0.26}$ & $\approx$ 300 sec \\[2pt]
\hline
FVBM   		& $\textbf{7.60} \pm \textbf{0.35}$  & $ \textbf{29.61} \pm \textbf{0.43} $ &  $\approx$ 1200 sec & $ 14.38 \pm 0.12 $ & $ \textbf{27.48} \pm \textbf{0.34} $ & $\approx$ 400 sec \\[2pt]
\hline
DAE			& $5.41 \pm 0.18$  & $23.80 \pm 0.43$ & $\approx$ 900 sec & $13.13 \pm 0.26 $ & $25.04 \pm 0.32$ & $\approx$ 400 sec \\[2pt]
\hline
\textbf{DEM}& $\textbf{7.77} \pm \textbf{0.19}$ $^{\alpha\beta\delta}$  & $\textbf{30.01} \pm  \textbf{0.86}$ $^{\alpha\beta\delta}$ & $\approx$ 1600 sec & $\textbf{15.49} \pm \textbf{0.23}$ $^{\alpha\beta\delta}$ & $\textbf{28.45} \pm \textbf{1.47}$ $^{\alpha\delta}$ & $\approx$ 400 sec \\[3pt]
\hline
\end{tabular}
\end{sc}
\end{center}
\end{table*}

In Table \ref{topkratingRecommendation_crossfolder_large}, it shows the experiment results on MovieLen10M (full) and Retail datasets; As we could see in the table \ref{topkratingRecommendation_crossfolder_large}, \textbf{DEM} could achieve significant better results than all the baseline methods except \textbf{FVBM} on MovieLen10M dataset; On retail dataset, \textbf{DEM} could outperforms all other baselines except \textbf{LBL}. Compared with other baselines, \textbf{LBL} and \textbf{FVBM} could obtain relative stable results on all the four dataset. It could show that Bayesian Bi-Dependence models could largely approximate to true data distribution in some real word applications. However, \textbf{DEM} could consistently outperform both \textbf{FVBM} and \textbf{LBL} shows that by incorporating the high-order dependence terms, \textbf{DEM} could typically achieve better results.

\begin{table*}[t]
\caption{Top@1 and Top@10 Prediction Accuracy on Epinions (75,879) and Slashdot(77,360)  Dataset. Superscripts $\alpha$, $\beta$, $\gamma$ and $\delta$ indicate statistically significant improvements $(p < 0.01)$ over DAE, FVBM, LBL and RBM}
\label{topkratingRecommendation_crossfolder_large_socialnetwork}
\begin{center}
\begin{sc}
\begin{tabular}{c|lll|lll}
\hline
\multicolumn{1}{c}{} & & & & \\[\dimexpr-\normalbaselineskip-\arrayrulewidth]%

\textbf{Models} & \multicolumn{3}{c|}{Epinions (75,879 Users)} & \multicolumn{3}{c}{Slashdot (77,360 Users)} \\
				 	& Top@1 	& Top@10	& Run Time	&	Top@1	& 	Top@10 & Run Time	\\ 
\hline 
CVG   		& $4.04 \pm 0.45 $  & $12.76 \pm 0.44$	& $\approx$ 30 sec & $2.12 \pm 0.22$ & $6.80 \pm 0.35 $ & $\approx$ 30 sec \\ [2pt]
\hline 
NormCVG   	& $5.01 \pm 0.50 $  & $15.53 \pm 0.57$ &  $\approx$ 30 sec  & $2.65 \pm 0.21$ & $7.64 \pm 0.38 $ & $\approx$ 30 sec \\[2pt]
\hline
LRW   		& $5.17 \pm 0.43 $  & $15.91 \pm 0.51$ & $\approx$ 1500 sec & $2.62 \pm 0.23$ & $7.65 \pm 0.41 $ & $\approx$ 2000 sec \\[2pt]
\hline
LDA    	    & $1.41 \pm 0.07$ & $6.50 \pm 0.61$ & $\approx$ 5100 sec & $0.96 \pm 0.14$ & $4.49 \pm 0.52 $ &  $\approx$ 8500 sec \\[2pt]
\hline
RBM		    & $2.45 \pm 0.15$  & $11.49 \pm 0.59$ & $\approx$ 3300 sec & $1.81 \pm 0.24 $ & $5.98 \pm 0.57$ & $\approx$ 5800 sec \\[2pt]
\hline
LBL         & $3.82 \pm 0.24$  & $12.72 \pm 0.60$ & $\approx$ 1500 sec & $2.39 \pm 0.39$ &  $6.77 \pm 0.18$ & $\approx$ 2200 sec \\[2pt]
\hline
FVBM   		& $2.93 \pm 0.31$  & $ 12.79 \pm 0.39 $ &  $\approx$ 3500 sec & $ 2.10 \pm 0.41 $ & $ 6.84 \pm 0.25 $ & $\approx$ 4300 sec \\[2pt]
\hline
DAE			& $5.29 \pm 0.44$  & $17.13 \pm 0.62$ & $\approx$ 3200 sec & $3.32 \pm 0.53 $ & $9.18 \pm 0.21$ & $\approx$ 5800 sec \\[2pt]
\hline
\textbf{DEM}& $\textbf{6.09} \pm \textbf{0.44}$ $^{\alpha\beta\gamma\delta}$  & $\textbf{18.83} \pm  \textbf{0.56}$ $^{\alpha\beta\gamma\delta}$ & $\approx$ 3500 sec & $\textbf{3.75} \pm \textbf{0.15}$ $^{\alpha\beta\gamma\delta}$ & $\textbf{9.42} \pm \textbf{0.49}$ $^{\alpha\beta\gamma\delta}$ & $\approx$ 4400 sec \\[3pt]
\hline
\end{tabular}
\end{sc}
\end{center}
\end{table*}

In Table \ref{topkratingRecommendation_crossfolder_large_socialnetwork}, we compared all the nine approaches on social network datasets. From the table \ref{topkratingRecommendation_crossfolder_large_socialnetwork}, it shows that heuristic methods \textbf{CVG}, \textbf{NormCVG} and \textbf{LRW} outperforms most of the advanced algorithms on Epinions and Slashdot dataset. One possible reason may due to the number of tail users in social networks is much larger than online Rating dataset and Product Co-Purchasing dataset. Only \textbf{DAE} and \textbf{DEM} models could outperform heuristic based models on Epinions and Slashdot dataset, while \textbf{DEM} significantly beat all baselines.

An interesting result from experiments is that the $L_k$ dependence model \textbf{RBM} does not perform better than $L_2$ dependence models (i.e., \textbf{LBL} and \textbf{FVBM}). There would be many factors to affect the model performance on different datasets, i.e., local optimization algorithm, hyperparameter selection, etc. Almost all the statistic models are biased towards/against some data distribution. For the experiment datasets in multiple domains, it is impractical to assume data generated from single distribution assumption. Therefore, in \textbf{DEM}, it proposed an from bottom to up schema to gradually learn the data distribution from low-order dependence assumptions to high-order dependence assumptions.

\subsection{An Analysis of Model Hyperparameters}

\begin{figure}
\begin{center}
\centerline{\includegraphics[width=3in]{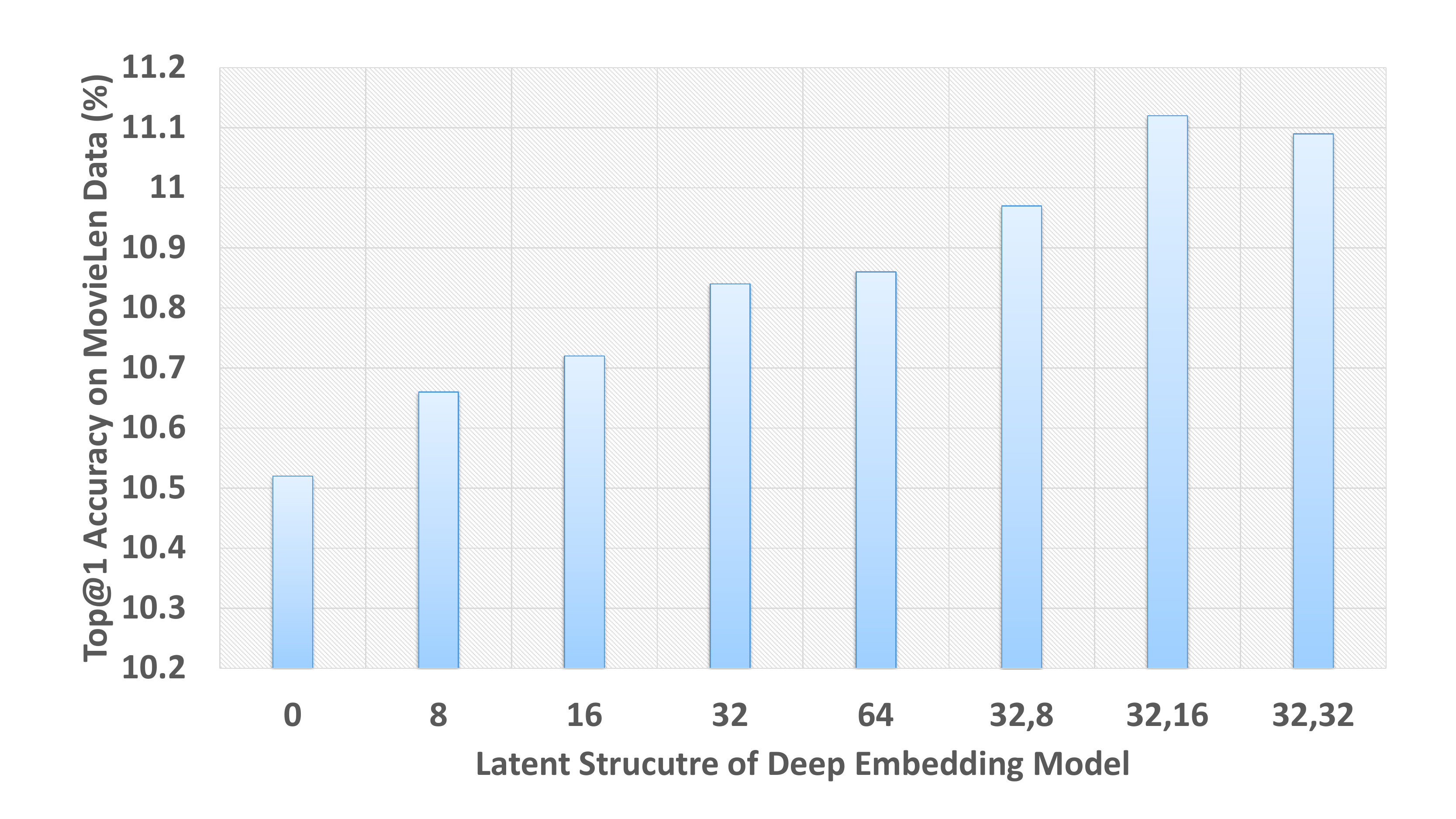}}
\caption{Hyper-Parameters Selection for Deep Embedding Model on MovieLen1M Data}
\label{movielenhyperparam}
\end{center}
\end{figure} 

In the subsection, we empirically analysis the hyperparameters in \textbf{DEM}. We take MovieLen1M dataset for experiment to show that how the model performance varied by selecting different model hyperparameters. In the Figure \ref{movielenhyperparam}, we compare the results of Top1 accuracy on different hyperparameter settings; \textbf{DEM-0} indicates the deep embedding model with no hidden layers.  \textbf{DEM-0} is equals to \textbf{FVBM}. \textbf{DEM-8}, \textbf{DEM-16}, \textbf{DEM-32} and \textbf{DEM-64} indicate the model has single hidden layer, with number of hidden states be $8$, $16$, $32$, and $64$ respectively. Likewise, \textbf{DEM-32 $\times$ 16} indicate the model contains two hidden layers with $32$ and $26$ hidden states at each layer respectively.  From the Figure \ref{movielenhyperparam}, we see the \textbf{DEM-32 $\times$ 16} could achieve the best performance compared with other hyperparameter settings of \textbf{DEM}.  However, the improvement of \textbf{DEM-32 $\times$ 16} over the other models is not significant.

\subsection{Learning Representations}
\begin{figure*}
		\centering
		\centerline{
			\subfigure[]{	
				\includegraphics[width=7.0in,height=5in]{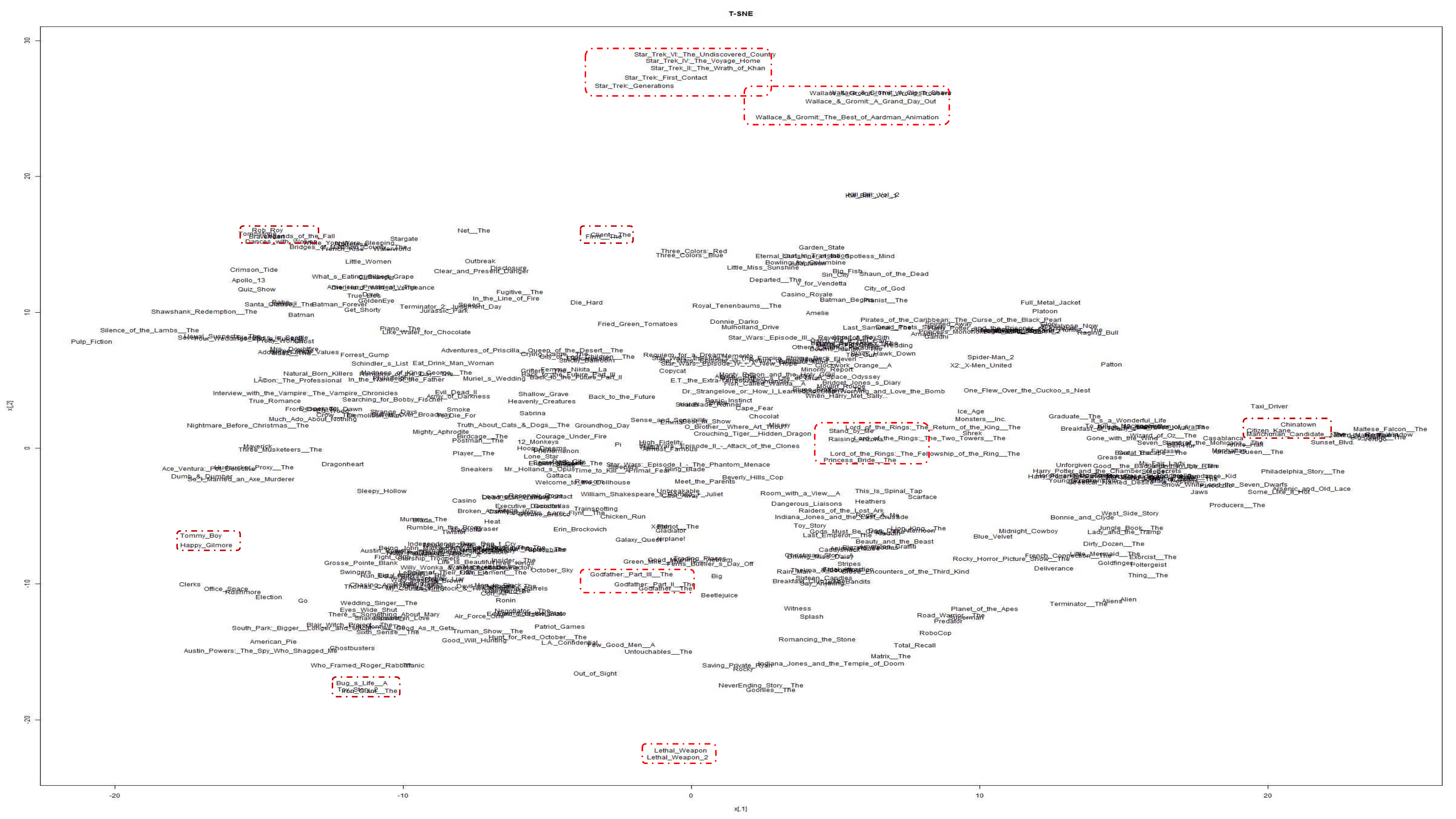}
			}	
		}
		\centerline{
			\subfigure[]{
				\includegraphics[width=6.0in]
				{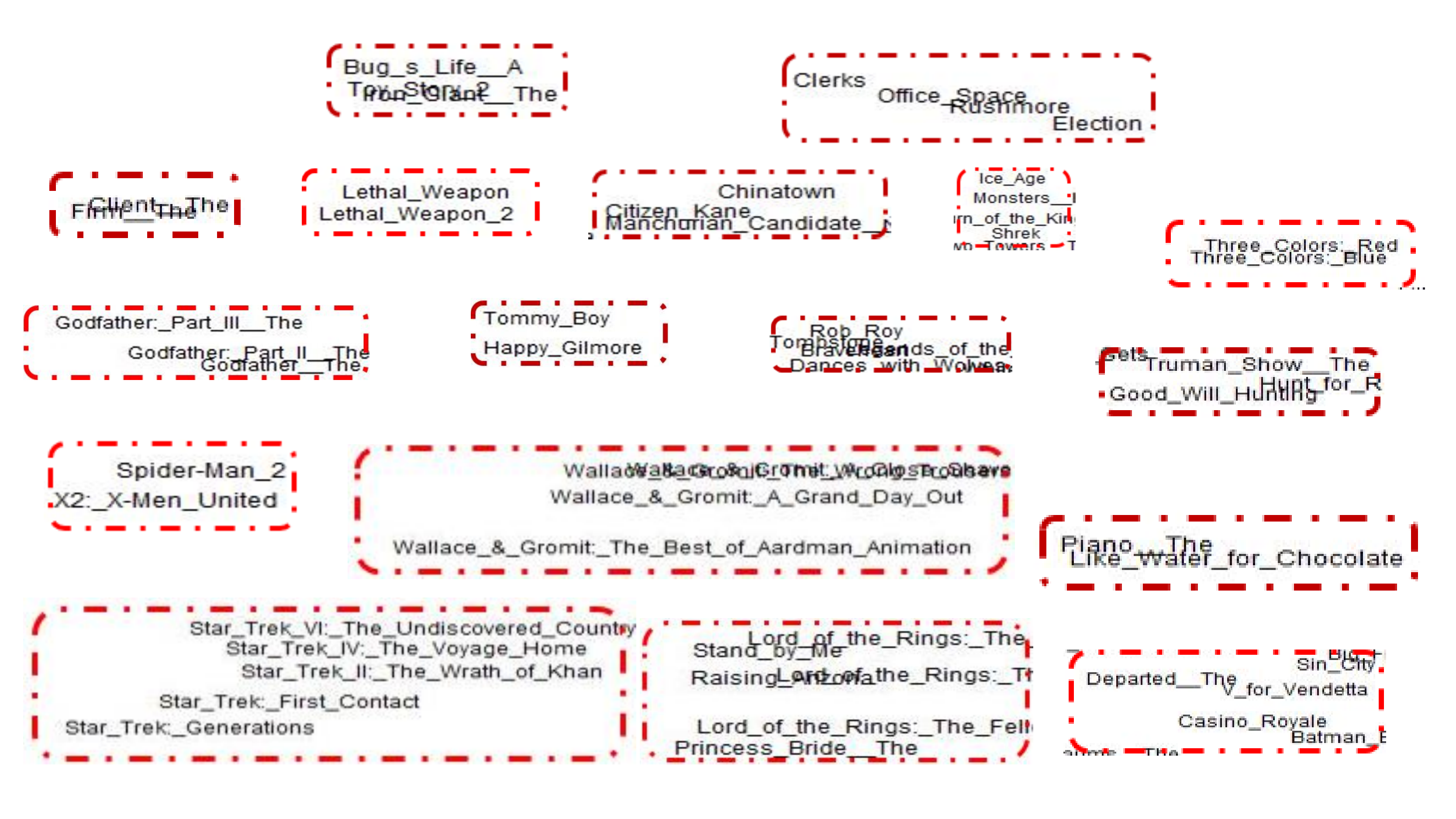}
			}	
		}
		\caption{DEM*-64 $\times$ 64  for 500 Movies Visualization on MovieLen1M dataset. (T-SNE visualization tool)}
		
		\label{bigvis}
	\end{figure*}

The DEM provides an unified framework which could joint train the $L_1$ dependence (Bias) term, $L_2$ dependence term and $L_k$ dependence (hierarchical latent embedding) term together for dynamic energy function estimation. In the deep learning area, it proposed that items' hidden semantic representations could be extracted from un-supervision data signal \cite{mnih2013learning}, i.e., co-occurrence data. The extracted latent semantic vectors could be able to used as semantic features for item classification and clustering tasks in the further. Therefore, in order to make the \textbf{DEM} learn the item semantic vector from co-occurrence data, we disable the $L_1$ and $L_2$ dependence terms, only keep the hidden neural layers. In the Table \ref{topkmoviehyper}, we present the experiment results of \textbf{DEM*} \footnote{\textbf{DEM*} is an simplified \textbf{DEM} which removes the $L_1$, $L_2$ terms in the dynamic energy function.} with different architectures. Among them, $\textbf{DEM*-64 $\times$ 64 }$ achieves best result, which obtains $11.02$ Top@1 Accuracy. It approximates to the best result $11.32$ by $\textbf{DEM-32 $\times$ 16 }$. We concatenate the hierarchical structured embedding vectors: $\{R^1_t, R^2_t, ..., R^k_t\}$ as an single semantic vector $R_t$ to represent the item $t$. 
In the model $\textbf{DEM*-64 $\times$ 64 }$, we obtain the $128$ dimension semantic vector for each movie. By projecting the $128$ dimension vectors into $2D$ image \footnote{We use the T-SNE \cite{van2008visualizing} visualization tool to obtain the movie visualization Figure.}, we obtain the movie visualization map as in Figure \ref{bigvis}. In the Figure \ref{bigvis}, it contains $500$ most frequent movies. As we can see in the figure, the distance between similar movies is usually closer than un-similar movies. We also give some informative pieces of movies in the graph. There are several movie series could be discovered and grouped together, i.e., \emph{Star Trek Series}, \emph{Wallace and Gromit Series} etc. 

\begin{table}[t]
\caption{Top@1 and Top10 Accuracy on MovieLen Data Set}
\label{topkmoviehyper}
\begin{center}
\begin{sc}
\begin{tabular}{l|cc}
\hline
\multicolumn{1}{c}{} & &\\[\dimexpr-\normalbaselineskip-\arrayrulewidth]%

\textbf{Models} & \multicolumn{2}{c}{MovieLen10M (Top500)} \\
				 	& Top@1 	& Top@10	\\
\hline

\textbf{DEM*-16}& $1.93 \pm 0.15 $ & $16.65 \pm 0.26$ \\
\textbf{DEM*-16 $\times$ 16 }& $9.11 \pm 0.12$ & $36.74 \pm 0.16$ \\
\textbf{DEM*-16 $\times$ 16 $\times$ 16 }& $9.95 \pm 0.43$ & $38.65 \pm 0.54$ \\
\textbf{DEM*-32}& $9.49 \pm 0.31$ & $38.91 \pm 0.36$ \\
\textbf{DEM*-32 $\times$ 32 }&   $10.26 \pm 0.18$ & $40.75 \pm 0.14$ \\
\textbf{DEM*-32 $\times$ 32 $\times$ 32 }&$10.51 \pm 0.31$ & $40.91 \pm 0.56$\\
\textbf{DEM*-64}& $10.35 \pm 0.28$ &  $40.93 \pm 0.41$ \\
\textbf{DEM*-64 $\times$ 64 }& $11.02 \pm 0.39$ & $41.28 \pm 0.39$ \\
\textbf{DEM*-64 $\times$ 64 $\times$ 64 }& $10.53 \pm 0.19$ & $40.84 \pm 0.18$ \\

\hline
\end{tabular}
\end{sc}
\end{center}
\end{table}

\section{Conclusion and Future Work}
In the paper, we introduce a general Bayesian framework for co-occurrence data modeling. Based on the framework, several previous machine learning models, i.e., Fully Visible Boltzmann Machine, Restricted Boltzmann Machine, Maximum Margin Matrix Factorization etc are studied, which could can be interpreted as one of three categories according to the $L_1$, $L_2$ and $L_k$ assumptions. As motivated by three Bayesian dependence assumptions, we developed a hierarchical structured model or DEM. The DEM is a unified model which combines both the low-order and high-order item dependence features. While the low-order item dependence features are captured at the bottom layer, and high-order dependence features are captured at the top layer. The experiments demonstrate the effectiveness of \textbf{DEM}. It outperforms baseline methods significantly on several public datasets. In the future work, we plan to further our study along the following directions: 1) to develop an nonparametric bayesian model to automatically infer the deep structure from data efficiently to avoid/reduce expensive hyper-parameter sweeping? 2) to develop an online algorithm to learn \textbf{DEM} on streaming co-occurrence data. 3) to encode the frequent item set information using the DEM representation?  








%

\bibliographystyle{abbrv}
\bibliography{sigproc}

\begin{thebibliography}{10}

\bibitem{bengio1999modeling}
Y.~Bengio and S.~Bengio.
\newblock Modeling high-dimensional discrete data with multi-layer neural
  networks.
\newblock In {\em NIPS}, volume~99, pages 400--406, 1999.

\bibitem{bengio2013generalized}
Y.~Bengio, L.~Yao, G.~Alain, and P.~Vincent.
\newblock Generalized denoising auto-encoders as generative models.
\newblock In {\em Advances in Neural Information Processing Systems}, pages
  899--907, 2013.

\bibitem{blei2003latent}
D.~M. Blei, A.~Y. Ng, and M.~I. Jordan.
\newblock Latent dirichlet allocation.
\newblock {\em the Journal of machine Learning research}, 3:993--1022, 2003.

\bibitem{das2007google}
A.~S. Das, M.~Datar, A.~Garg, and S.~Rajaram.
\newblock Google news personalization: scalable online collaborative filtering.
\newblock In {\em Proceedings of the 16th international conference on World
  Wide Web}, pages 271--280. ACM, 2007.

\bibitem{frey1998graphical}
B.~J. Frey.
\newblock {\em Graphical models for machine learning and digital
  communication}.
\newblock MIT press, 1998.

\bibitem{geman1986markov}
S.~Geman and C.~Graffigne.
\newblock Markov random field image models and their applications to computer
  vision.
\newblock In {\em Proceedings of the International Congress of Mathematicians},
  volume~1, page~2, 1986.

\bibitem{Globerson:2007:EEC:1314498.1314572}
A.~Globerson, G.~Chechik, F.~Pereira, and N.~Tishby.
\newblock Euclidean embedding of co-occurrence data.
\newblock {\em J. Mach. Learn. Res.}, 8:2265--2295, Dec. 2007.

\bibitem{gong1981pseudo}
G.~Gong and F.~J. Samaniego.
\newblock Pseudo maximum likelihood estimation: theory and applications.
\newblock {\em The Annals of Statistics}, pages 861--869, 1981.

\bibitem{hinton2006reducing}
G.~E. Hinton and R.~R. Salakhutdinov.
\newblock Reducing the dimensionality of data with neural networks.
\newblock {\em Science}, 313(5786):504--507, 2006.

\bibitem{hinton1986learning}
G.~E. Hinton and T.~J. Sejnowski.
\newblock Learning and relearning in boltzmann machines.
\newblock {\em MIT Press, Cambridge, Mass}, 1:282--317, 1986.

\bibitem{Hofmann:1998:SMC:888741}
T.~Hofmann and J.~Puzicha.
\newblock Statistical models for co-occurrence data.
\newblock Technical report, Cambridge, MA, USA, 1998.

\bibitem{hyvarinen2006consistency}
A.~Hyv{\"a}rinen.
\newblock Consistency of pseudolikelihood estimation of fully visible boltzmann
  machines.
\newblock {\em Neural Computation}, 18(10):2283--2292, 2006.

\bibitem{johansen1990maximum}
S.~Johansen and K.~Juselius.
\newblock Maximum likelihood estimation and inference on cointegration—with
  applications to the demand for money.
\newblock {\em Oxford Bulletin of Economics and statistics}, 52(2):169--210,
  1990.

\bibitem{larochelle2010tractable}
H.~Larochelle, Y.~Bengio, and J.~Turian.
\newblock Tractable multivariate binary density estimation and the restricted
  boltzmann forest.
\newblock {\em Neural computation}, 22(9):2285--2307, 2010.

\bibitem{larochelle2011neural}
H.~Larochelle and I.~Murray.
\newblock The neural autoregressive distribution estimator.
\newblock {\em Journal of Machine Learning Research}, 15:29--37, 2011.

\bibitem{le2008representational}
N.~Le~Roux and Y.~Bengio.
\newblock Representational power of restricted boltzmann machines and deep
  belief networks.
\newblock {\em Neural Computation}, 20(6):1631--1649, 2008.

\bibitem{maas2010probabilistic}
A.~L. Maas and A.~Y. Ng.
\newblock A probabilistic model for semantic word vectors.
\newblock In {\em NIPS Workshop on Deep Learning and Unsupervised Feature
  Learning}, 2010.

\bibitem{mnih2007three}
A.~Mnih and G.~Hinton.
\newblock Three new graphical models for statistical language modelling.
\newblock In {\em Proceedings of the 24th international conference on Machine
  learning}, pages 641--648. ACM, 2007.

\bibitem{mnih2013learning}
A.~Mnih and K.~Kavukcuoglu.
\newblock Learning word embeddings efficiently with noise-contrastive
  estimation.
\newblock In {\em Advances in Neural Information Processing Systems}, pages
  2265--2273, 2013.

\bibitem{Mordohai:2010:DEM:1756006.1756018}
P.~Mordohai and G.~Medioni.
\newblock Dimensionality estimation, manifold learning and function
  approximation using tensor voting.
\newblock {\em J. Mach. Learn. Res.}, 11:411--450, Mar. 2010.

\bibitem{ravikumar2010high}
P.~Ravikumar, M.~J. Wainwright, J.~D. Lafferty, et~al.
\newblock High-dimensional ising model selection using ℓ1-regularized
  logistic regression.
\newblock {\em The Annals of Statistics}, 38(3):1287--1319, 2010.

\bibitem{salakhutdinov2007restricted}
R.~Salakhutdinov, A.~Mnih, and G.~Hinton.
\newblock Restricted boltzmann machines for collaborative filtering.
\newblock In {\em Proceedings of the 24th international conference on Machine
  learning}, pages 791--798. ACM, 2007.

\bibitem{sohl2011minimum}
J.~Sohl-dickstein, P.~Battaglino, and M.~R. Deweese.
\newblock Minimum probability flow learning.
\newblock In {\em Proceedings of the 28th International Conference on Machine
  Learning (ICML 2011)}, pages 905--912, 2011.

\bibitem{srebro2004maximum}
N.~Srebro, J.~Rennie, and T.~S. Jaakkola.
\newblock Maximum-margin matrix factorization.
\newblock In {\em Advances in neural information processing systems}, pages
  1329--1336, 2004.

\bibitem{swersky2011autoencoders}
K.~Swersky, D.~Buchman, N.~D. Freitas, B.~M. Marlin, et~al.
\newblock On autoencoders and score matching for energy based models.
\newblock In {\em Proceedings of the 28th International Conference on Machine
  Learning (ICML-11)}, pages 1201--1208, 2011.

\bibitem{van2008visualizing}
L.~Van~der Maaten and G.~Hinton.
\newblock Visualizing data using t-sne.
\newblock {\em Journal of Machine Learning Research}, 9(2579-2605):85, 2008.

\bibitem{vincent2011connection}
P.~Vincent.
\newblock A connection between score matching and denoising autoencoders.
\newblock {\em Neural computation}, 23(7):1661--1674, 2011.

\bibitem{vincent2008extracting}
P.~Vincent, H.~Larochelle, Y.~Bengio, and P.-A. Manzagol.
\newblock Extracting and composing robust features with denoising autoencoders.
\newblock In {\em Proceedings of the 25th international conference on Machine
  learning}, pages 1096--1103. ACM, 2008.

\bibitem{yildirim2008random}
H.~Yildirim and M.~S. Krishnamoorthy.
\newblock A random walk method for alleviating the sparsity problem in
  collaborative filtering.
\newblock In {\em Proceedings of the 2008 ACM conference on Recommender
  systems}, pages 131--138. ACM, 2008.

\bibitem{zhang2007binary}
Z.~Zhang, C.~Ding, T.~Li, and X.~Zhang.
\newblock Binary matrix factorization with applications.
\newblock In {\em Data Mining, 2007. ICDM 2007. Seventh IEEE International
  Conference on}, pages 391--400. IEEE, 2007.

\end{thebibliography}

\end{document}